\documentclass[lettersize,journal]{IEEEtran}
\usepackage{amsmath,amsfonts}
\usepackage{algorithmic}
\usepackage{algorithm}
\usepackage{array}
\usepackage[caption=false,font=normalsize,labelfont=sf,textfont=sf]{subfig}
\usepackage{textcomp}
\usepackage{stfloats}
\usepackage{url}
\usepackage{verbatim}
\usepackage{graphicx}

\usepackage{graphicx}
\usepackage{enumerate}
\usepackage{xcolor}
\usepackage{url} 
\usepackage{tabularx}
\usepackage[square, numbers]{natbib}
\usepackage{etoolbox} 
\makeatletter
\renewcommand{\citet}[1]{%
    \citeauthor{#1}~(\citeyear{#1})}
\makeatother

\usepackage[utf8]{inputenc}
\usepackage[T1]{fontenc}
\usepackage{multirow}

\DeclareMathOperator{\var}{Var}
\newtheorem{definition}{Definition}

\begin{document}

\title{Order-based Causal Discovery for Multistage Processes}

\author{Eun-Yeol~Ma, Junsub~Jung, and~Heeyoung~Kim
\thanks{All authors are with the Department of Industrial and Systems Engineering, Korea Advanced Institute of Science and Technology, Daejeon 34141, South Korea (e-mail: heeyoungkim@kaist.ac.kr).}
\thanks{}}

\markboth{IEEE Transactions on Knowledge and Data Engineering}%
{Shell \MakeLowercase{\textit{et al.}}: A Sample Article Using IEEEtran.cls for IEEE Journals}
\maketitle

\begin{abstract}
Causality has become an increasingly important tool for gaining a deeper understanding of complex systems. Among various causal analysis methods, causal discovery, which identifies causal relationships among variables from data, has been widely used to uncover underlying causality in diverse processes. However, while multistage processes are prevalent in many fields, existing causal discovery methods may produce counterintuitive results, given the known process knowledge, and may not be computationally efficient for handling large datasets typical of multistage processes. 
To address this gap, we propose a novel causal discovery method called Order-based Causal Discovery for Multistage Processes (OCDM). OCDM is designed to infer the causal structure of multistage data while preserving their inherent hierarchical and sequential structure by explicitly incorporating process knowledge into the causal discovery process. Specifically, we propose a structural knowledge-informed order-inferring algorithm that infers the causal order of variables by incorporating information about the stage from which each variable originates, based on an order-based causal discovery framework naturally suited for inherently ordered multistage data. Furthermore, to eliminate spurious edges from the initial causal graph generated based on the inferred causal order, we introduce a novel pruning technique using stochastic gated neural networks, which offers greater computational efficiency compared to existing methods. Through experiments on various datasets, we demonstrate that OCDM effectively infers the causal structure of multistage processes, outperforming existing methods.

\end{abstract}


\section{Introduction}
\IEEEPARstart{M}{ultistage} processes are prevalent across a wide range of fields, including manufacturing processes, large-scale information systems, and personalized service operations \citep{shi2009quality}.  
As processes within various fields become increasingly complex, it is challenging even for domain experts to fully understand entire systems based solely on existing domain knowledge. Consequently, data-driven methodologies, such as machine learning and deep learning, have been actively used to enhance a wide range of tasks. However, most of the methods rely on associative relationships between variables, which can lead to misunderstandings and, ultimately, inappropriate decision-making. In contrast, understanding causality and the underlying causal structure of a system enables us to anticipate the effects of changes, leading to more precise and robust decision-making \citep{peters2017elements, ma2024simultaneous}.

Among various causal modeling techniques, causal discovery (or causal structure learning) identifies the causal relationships among variables in a system based on data \citep{vowels2022d}. Causal discovery algorithms aim to learn the direct causes of each variable in the dataset. Under the causal Markov assumption \citep{pearl2009causality}, these  relationships can be represented as structural causal models (SCM), where each variable is expressed as a function of its direct causes or parents, along with some exogenous noise. Therefore, learning the causal structure can be seen as recovering the underlying data-generating process. In addition, all SCMs can be translated into a directed acyclic graph (DAG), where each variable is represented as a node and a directed edge from one node to another indicates a parent-child relationship between the variables. 
Throughout the paper, we interchangeably use probabilistic terminology and graphical terminology. The inferred causal structure can then be leveraged for a variety of downstream tasks that require causal reasoning, including root cause analysis, fault detection and diagnosis, process optimization, and post-intervention modeling \citep{budhathoki2022causal,kim2023contextual,soh2018application,wang2024fault,cho2023prediction,budhathoki2021did}. 

However, existing causal discovery methods are not well-suited for multistage processes. First, conventional causal discovery algorithms may produce counterintuitive results, given the known process knowledge about the stage from which a variable is collected. Second, while multistage data are often large, with the number of variables growing with the number of stages, these algorithms are prone to the curse of dimensionality, as the search space of DAGs expands super-exponentially with the number of nodes \citep{teyssier2005ordering}.

To address theses issues, we propose a novel causal discovery method called Order-based Causal Discovery for Multistage Processes (OCDM), designed to infer the causal structure of multistage data while preserving their inherent hierarchical and sequential structure, and improving computational efficiency compared to conventional algorithms. First, to ensure that OCDM maintains the inherent structure of multistage data, we propose a structural knowledge-informed order-inferring algorithm. This algorithm explicitly incorporates process knowledge about the stage from which each variable originates and is built on an order-based causal discovery framework \citep{rolland2022score, sanchez2023diffusion} to maximize the use of information related to the ordered nature of multistage processes.Specifically, the process of searching for the causal order occurs stage-wise, with the order of nodes within each stage fully  determined before moving on to identify the order of nodes in the next stage. This approach ensures the order of stages is maintained in the resultant order of nodes. The inferred causal order can then be used to generate a DAG by connecting all nodes that succeed a given node in the inferred order. However, this initial DAG is likely to contain spurious edges, which must be removed. Previous order-based causal discovery methods typically utilize the sparse spline regression-based pruning technique proposed by \citet{buhlmann2014cam}, which may struggle with high-dimensional data containing a large number of samples---common in multistage process datasets. As a more suitable alternative, we propose a novel pruning method based on stochastic gated neural networks (STG-NN) \citep{yamada2020feature}, leveraging the flexibility and predictive power of $L_0$-regularized neural networks. We are the first to use neural networks in the pruning phase of order-based causal discovery, aiming to develop a more effective pruning method tailored to the unique characteristics of multistage data.
We evaluated the proposed OCDM on simulated multistage graphs, as well as on pseudo-real data that models a real multistage process, demonstrating its superiority over previous causal discovery methods in learning the causal structure of multistage processes.

\section{Related works}
The majority of previously suggested causal discovery methods can be classified as either constraint-based or score-based \citep{vowels2022d}. Constraint-based methods (e.g., PC-algorithm \citep{spirtes1991algorithm}) rely on repeated conditional-independence tests to infer the causal structure of the data. Because these tests must be conducted for every possible pair of variables, conditioned on all possible sets of other variables, the number of tests increases exponentially. Furthermore, the accuracy of these tests declines rapidly as the size of the conditioning set increases \citep{li2020nonparametric}, limiting their applicability to low-dimensional settings, which are rare in multistage processes. In contrast, score-based methods infer the causal structure by optimizing a certain metric (e.g., likelihood). Traditionally, this optimization is performed in a greedy manner (e.g., GES \citep{chickering2002optimal}), which is computationally expensive. To address these drawbacks, a line of score-based methods has been proposed that translate the combinatorial search into a continuous optimization process (e.g., NOTEARS \citep{zheng2018dags}). These methods, including those using neural networks, optimize a target metric with an acyclicity constraint to learn a causal DAG, improving computational efficiency compared to traditional combinatorial methods \citep{lachapelle2019gradient, yu2019dag, zheng2020learning}. However, these methods may fail to return a valid causal structure when poorly trained, as the output is guaranteed to be a DAG only when the acyclicity constraint is zero. Furthermore, as the number of nodes increases, these methods can struggle to learn the causal structure due to the growing computational complexity of enforcing the acyclicity constraint.

To overcome the limitations of constraint-based and score-based methods, order-based causal discovery methods have been proposed. In order-based causal discovery, the causal order is defined as a permutation of the nodes in a given DAG, where a node appears in the order only after its parents \citep{peters2017elements} (see Supplemental Materials for the formal definition), and is inferred prior to inferring the causal DAG. Although searching for causal orders is inherently  combinatorial, inferring causal orders is a significantly more manageable task than inferring DAGs. Furthermore, order-based methods always ensure that the inferred causal structure is a valid DAG \citep{buhlmann2014cam}, meaning that no cycles exist in the inferred structure. This approach was first introduced by \citet{buhlmann2014cam} through the causal additive model (CAM), which infers the causal order under a nonlinear additive effects model with additive Gaussian noise. Assuming this causal model, any ordering that does not align with the ground-truth DAG leads to a lower expected likelihood than the true causal order, allowing the causal order to be learned through greedy maximum likelihood estimation. However, this greedy approach heavily burdens the causal order search process. To accelerate the causal order search, other methods such as SCORE \citep{rolland2022score}, DAS \citep{montagna2023scalable}, and DiffAN \citep{sanchez2023diffusion} have been proposed. These methods utilize the score function (gradient of the log-likelihood) instead of greedily maximizing the likelihood. However, similar to other conventional causal discovery methods, they typically do not incorporate available prior structural knowledge or merely include or exclude specific edges. 

Order-based causal discovery methods estimate the causal DAG by pruning, or removing spurious edges, from a DAG estimate generated using the causal order. \citet{buhlmann2014cam} proposed a sparse spline regression-based pruning procedure, where a spline regression model with group LASSO regularization \citep{ravikumar2009sparse} is constructed for each variable, taking all preceding variables in the inferred causal order as input. The subset of variables that minimizes the prediction loss is selected, and edges from unselected variables are pruned from the initial DAG to obtain the final estimate. This pruning procedure, typically employed by all order-based causal discovery methods, will be referred to as CAM-pruning hereafter \citep{buhlmann2014cam}. Despite its widespread use, CAM-pruning may be inadequate in terms of both pruning accuracy and computational efficiency when learning the causal structure of multistage processes, whose dimensionality is typically higher than that considered by previously proposed order-based causal discovery methods.

While the utilization of structural knowledge in order-based discovery has primarily been studied within homogeneous tabular settings, parallel methodological advances have recently emerged in cross-modal and multimedia domains. In fields such as vision-language and video reasoning, researchers increasingly leverage domain-specific structural priors to constrain causal graphs and eliminate spurious correlations \citep{liu2022causal, liu2023cross, wei2023visual, chen2025a, chen2025b}. For instance, mapping vision-text relations into structural causal frameworks allows models to ground questions or generate reports without relying on statistical biases. Mirroring this philosophy of leveraging macroscopic structural knowledge to reduce the search space of complex systems, our work focuses on exploiting the known sequential stage boundaries inherent in multistage processes to guide order-based causal discovery.

\section{Proposed method}
\subsection{Data-generating process}
Let $\mathbf{X} \in \mathbb{R}^{n \times d}$ denote the $d$-dimensional observed dataset of $n$ samples. We denote the parents (direct causes) of a variable $\mathbf{X}_i$ as $Pa(\mathbf{X}_i)$ and children (direct effects) as $Ch(\mathbf{X}_i)$. 
As commonly assumed in the literature \citep{lachapelle2019gradient, zheng2018dags, ng2022masked, rolland2022score, sanchez2023diffusion}, we assume that the data-generating process of the observed data follows a nonlinear additive noise model:
\begin{equation}
    \mathbf{X}_i = f_i(Pa(\mathbf{X}_i)) + U_i, \quad U_i \sim \mathcal{N}(0, \sigma_i^2), \quad i=1,...,d,
\label{eq:nonlinear_additive_noise}
\end{equation}
where $f_i$ is a nonlinear link function and $U_i$ represents independent exogenous Gaussian noise. We also assume causal sufficiency, meaning there are no unobserved confounders. This nonlinear additive noise model is identifiable from observational data \citep{hoyer2008nonlinear, peters2014causal}.

Furthermore, we assume that the data-generating process can be represented as a layered DAG \citep{healy2013hierarchical}, where edges can only exist between nodes within the same layer or from a node in an upper layer to a node in a lower layer (see Supplemental Materials for the formal definition). 
The data-generating process of a multistage process naturally forms a layered DAG, as an object is sequentially processed according to the order of stages. We assume that the stage from which each variable originated is known, but the causal relationships among variables---regardless of whether they originate from the same stage---remain unknown. This assumption is plausible for most multistage processes. 

\subsection{Order-based Causal Discovery for Multistage Processes (OCDM)}
We propose a new causal discovery method tailored for inferring the causal structure of multistage processes, called Order-based Causal Discovery for Multistage Processes (OCDM). This method leverages the multistage structure of the data by directly integrating the structural knowledge into the causal discovery algorithm. 

Specifically, OCDM first infers the causal order using a structural knowledge-informed order-inferring algorithm. Unlike previous methods that lack the necessary considerations for accurately inferring the causal order in multistage processes, OCDM incorporates available knowledge about the multistage structure and the stage from which each variable originated, preventing counterintuitive causal orders. This approach improves the quality of the causal order estimate. Once the causal order is estimated, it is used to generate an initial estimate of the true DAG by constructing a graph with edges connecting each node to all nodes that appear after it in the order. For clarity, we refer to this graph as the \emph{order-connected graph}, which is formally defined as follows. 
\begin{definition}[Order-connected graph]
Let a DAG $G$ be defined by the tuple of the vertex set $\mathcal{V}$ and edge set $\mathcal{E}$. Given a causal order $\pi$, an \textit{order-connected graph} $G^{\pi}=(\mathcal{V}, \mathcal{E}^{\pi})$ is a DAG defined such that, if $\pi_i < \pi_j$, then $(i,j)\in \mathcal{E}^{\pi}$.
\label{def:order-connected}
\end{definition}
The constructed order-connected graph is inherently a DAG, without requiring any additional measures to ensure acyclicity. Furthermore, any subgraph of the order-connected graph is also inherently a DAG \citep{buhlmann2014cam}.  The order-connected graph can then be pruned of spurious edges using the conventional CAM-pruning technique based on sparse regression \citep{buhlmann2014cam} to obtain the final DAG estimate. However, CAM-pruning can become inefficient and inaccurate when applied to high-dimensional data typical of multistage processes---contradicting the strengths of order-based causal discovery methods in efficiently handling such data \citep{rolland2022score}. 
To address this limitation, OCDM introduces a novel pruning method based on stochastic gated neural networks (STG-NN) \citep{yamada2020feature}, enhancing pruning performance in the context of multistage processes. The overall procedure of OCDM is summarized in Algorithm \ref{alg:ocdm}, with the order search and pruning strategies detailed in Section \ref{order} and Section \ref{prune}, respectively. 

\begin{algorithm}
\caption{Causal discovery via OCDM}
\label{alg:ocdm}
\begin{algorithmic}[1]
\REQUIRE Data $\mathbf{X} \in \mathbb{R}^{n \times d}$, node index $\mathcal{V}={L_1 \cup \dots \cup L_w}$, where $L_i$ is the set of $k_i$ nodes in stage $i$, batch size $b$
\ENSURE Causal order $\pi$, causal DAG $G^{\pi}_{*}$
\newline
\textbf{Notation:} Let $-\pi = \mathcal{V} \setminus \pi$.
\STATE Initialize the causal order $\pi = [ ]$,  Mask $\mathbf{M} = \mathbf{1}^{b \times d}$
\STATE Train the diffusion model $\boldsymbol{\epsilon}_\theta$
\FOR{$i = w, w-1, \ldots, 1$}
    \WHILE{$L_i \neq \emptyset$}
        \STATE $\mathbf{B} \xleftarrow{b} \mathbf{X}$;
        \STATE $\mathbf{B} \leftarrow \mathbf{B} \odot \mathbf{M}_{\pi}$;
        \STATE Estimate the score function \newline $s_{-\pi}(\mathbf{B}) = \nabla_{\mathbf{X}_{-\pi}} \log p(\mathbf{B})$;
        \STATE Estimate the variance of score's Jacobian \newline  $V_j = \text{Var}_{\mathbf{X}_{-\pi}}\left[\frac{\partial s_j(\mathbf{B})}{\partial x_j}\right]$;
        \STATE $l \leftarrow \arg\min_{j\in L_i} V_j$;
        \STATE $\pi\leftarrow [l, \pi]$;
        \STATE $L_i \leftarrow L_i \setminus \{l\}$;
        \STATE $\mathbf{M}_l = 0$;
    \ENDWHILE
\ENDFOR
\STATE Generate the order-connected graph $G^\pi=(\mathcal{V}, \mathcal{E}^{\pi})$;
\STATE Initialize $G^{\pi}_{*} \leftarrow G^{\pi}$;
\FOR{$i = 2$ to $d$}
    \STATE $Y_i \leftarrow \mathbf{X}_{\pi_i}$;
    \STATE Fit $\hat{Y}_i = f_i(\mathbf{X}_{\pi_{0:i-1}})$, where $f_i$ is an STG-NN
    \STATE $G^{\pi}_{*} \leftarrow (\mathcal{V}, \mathcal{E}^{\pi} \setminus \{(j, i) \mid z_j = 0, j \in \pi_{0:i-1}\})$ (Eq. \eqref{eq:stg});
\ENDFOR
\end{algorithmic}
\end{algorithm}

\subsection{Multistage-informed causal order search}\label{order}
Our approach to inferring the causal order and, subsequently, the order-connected graph is based on score matching, chosen for its theoretical guarantees and computational efficiency in inferring causal orders under less restrictive conditions \citep{rolland2022score}. Assuming that the given data were generated from a nonlinear additive Gaussian noise model (Eq. \eqref{eq:nonlinear_additive_noise}), its log-likelihood can be expressed as follows:
\begin{align}
    \log{p(\mathbf{X})} 
    &= \sum_{i=1}^{d} \log{p(\mathbf{X}_i \mid Pa(\mathbf{X}_i))} \notag \\
    &= -\frac{1}{2} \sum_{i=1}^{d} \left( \frac{\mathbf{X}_i - f_i(Pa(\mathbf{X}_i))}{\sigma_i} \right)^2 
    - \frac{1}{2} \sum_{i=1}^{d} \log{2\pi\sigma_i^2}.
\label{eq:likelihood}
\end{align}
Then, the score function $s_i(\mathbf{X}) \equiv \nabla_{x_i} \log{p(\mathbf{X})}$ can be expressed as follows:
\begin{align}
    s_i(\mathbf{X}) 
    &\equiv \nabla_{X_i} \log{p(\mathbf{X})} \notag \\
    &= -\frac{\mathbf{X}_i - f_i(Pa(\mathbf{X}_i))}{\sigma_i^2} \notag \\
    &\quad + \sum_{j \in Ch(\mathbf{X}_i)} \frac{\partial f_j}{\partial \mathbf{X}_i}(Pa(\mathbf{X}_j)) 
    \left( \frac{\mathbf{X}_j - f_j(Pa(\mathbf{X}_j))}{\sigma_j^2} \right).
\label{eq:score}
\end{align}
Because leaf nodes are childless by definition and a node cannot be its own parent in a DAG, $\frac{\partial s_i(\mathbf{X})}{\partial \mathbf{X}_i} = -\frac{1}{\sigma_i^2}$, which is constant, resulting in $\var{ \left[ \frac{\partial s_i(\mathbf{X})}{\partial \mathbf{X}_i} \right] } = 0$ \citep{rolland2022score}. Therefore, given a score estimator, leaf nodes can be identified iteratively by finding the node with the least variance in the Jacobian of the score to infer the full causal order. 

Building upon the score matching-based causal order search framework, OCDM incorporates prior knowledge about the multistage structure of the process to more accurately infer the causal order. The overall procedure is outlined in lines 1 to 14 of Algorithm \ref{alg:ocdm}. Specifically, OCDM constrains the search space for leaf nodes within each stage. Based on the variance in the Jacobian of the score, nodes are first ordered within their respective stages to form a stage-wise causal order, which is then combined to produce the complete causal order of the process. Because the causal order is learned from leaf to root, the order search begins at the last stage $L_{w}$. After determining the causal order of all nodes in the last stage, based on the score's Jacobian for the remaining unordered nodes, OCDM proceeds upwards to the preceding stage $L_{w-1}$ and infers its node order. This process continues until the nodes in the first stage $L_1$ are fully ordered.

The proposed order-inferring algorithm fully leverages the layered DAG structure of multistage processes, which leads to improved estimation accuracy. Because edges from lower to upper layers are prohibited in layered DAGs, a node in a lower layer can never precede a node in an upper layer in the causal order. OCDM enforces this structural constraint by performing stage-wise ordering, starting from the last stage and progressing upward. By using stage-wise information to establish a coarse order of nodes, the algorithm eliminates the possibility of counterintuitive causal orders, where downstream variables precede upstream ones, and prevents false edges from lower to upper nodes, thereby enhancing the accuracy of the inferred causal structure.

We use DiffAN \citep{sanchez2023diffusion} as our base score-matching-based causal order search method, although other order-based causal discovery methods can also be used. We expect any base method to exhibit similar improvements to those observed with DiffAN when the stage-wise causal order search scheme proposed by OCDM is incorporated, as the benefits of leveraging stage information for causal order learning are independent of the specific underlying order search method. DiffAN utilizes diffusion probabilistic models to estimate the score function and employs the ``deciduous score,'' which adjusts the score for previously ordered nodes without requiring the retraining of the neural network. Consistent with the score matching-based causal order search framework, DiffAN iteratively identifies the leaf node and masks it (i.e., removes it from the data by replacing it with 0) to obtain the causal order. Specifically, this masking mechanism utilizes a dynamic matrix $\mathbf{M} \in \{0, 1\}^{b \times d}$ initialized to all ones. At each iteration, the data batch $\mathbf{B}$ is element-wise multiplied by the mask ($\mathbf{B} \odot \mathbf{M}_{\pi}$). When a node $l$ is identified as a leaf and added to $\pi$, its mask entries are set to zero ($\mathbf{M}_l = 0$). This isolates the remaining unordered variables ($\mathbf{X}_{-\pi}$), forcing the network to compute gradients strictly for active features. Consequently, OCDM leverages DiffAN's strong performance in score estimation and computational efficiency, while improving the quality of the inferred causal order through its novel search algorithm that incorporates stage information.

The computational complexity of the multistage-informed causal order search in OCDM directly follows that of its baseline method, DiffAN, which is $\mathcal{O}(n+d^3)$ \citep{sanchez2023diffusion}, where $n$ is the number of training samples and $d$ is the number of variables. Training the diffusion model $\boldsymbol{\epsilon}_\theta$ scales linearly with the number of samples ($\mathcal{O}(n)$), while determining the causal order has a complexity of $\mathcal{O}(d \cdot (d-1) \cdot i)$ with $i$ varying from 0 to $d$. The stage-wise search in OCDM only restricts the \emph{order search} to candidate leaf nodes within the stage under consideration. Because this does not alter the training scheme, the computational complexity remains identical to that of the base method.

\subsection{Stochastic gated neural network pruning}\label{prune}
The final estimate of the causal structure can be obtained by removing spurious edges from the order-connected graph (Def. \ref{def:order-connected}) generated using the inferred causal order.
While multistage processes are typically high-dimensional, pruning DAGs for high-dimensional data is considerably more challenging than for lower-dimensional data, because the complexity of pruning candidate edges for nodes further down the causal order increases as the dimensionality of the initial input increases.
To address this, OCDM considers the characteristics of multistage processes not only during causal order inference but also during pruning. Specifically, we propose a new pruning method based on neural networks with variable selection capabilities, STG-NN \citep{yamada2020feature}. We argue that the proposed approach is more suitable for multistage process data compared to the conventional CAM-pruning \citep{buhlmann2014cam} used in previous order-based causal discovery methods. CAM-pruning may struggle to accurately identify the true causal parents from a large number of candidate variables as the dataset grows in both dimensionality and sample size. 

The primary advantage of STG-NN pruning over CAM-pruning lies in its greater modeling flexibility and stronger alignment with the assumed causal model. Specifically, STG-NN pruning is consistent with the causal model assumed in this study (Eq. \eqref{eq:nonlinear_additive_noise}), whereas CAM-pruning implicitly assumes a causal model with additive effects through its use of spline regression for variable selection \citep{buhlmann2014cam}. This imposes an unnecessary constraint on the causal model that is not present during causal order inference. The additive structure of CAM-pruning inherently limits its ability to capture high-order causal interactions, particularly as the input dimensionality increases. In contrast, STG-NN pruning employs neural networks as the base nonlinear regression model. Their holistic, end-to-end training offers greater flexibility in modeling complex nonlinear functions and interactions among input variables, making STG-NN pruning more capable of discovering true causal relationships. Furthermore, STG-NN pruning utilizes a relaxed $L_0$-regularization approach for variable selection \citep{yamada2020feature}, whereas CAM-pruning relies on $L_1$-regularization. Although both enable variable selection, $L_0$-regularization directly constrains the number of selected variables and thus is inherently more suitable for this task, whereas $L_1$-regularization often introduces unwanted shrinkage effects \citep{tibshirani1996regression}.

STG-NN pruning removes spurious edges from the initial order-connected graph by training a neural network for each node. As outlined in lines 15 to 20 of Algorithm \ref{alg:ocdm}, these neural networks are trained sequentially for each target variable $Y_i = \mathbf{X}_{\pi_i}$ to predict its value using all preceding nodes in the causal order ($\mathbf{X}_{\pi_{0:i-1}}$) as the input vector. As a result, the input for each neural network includes both true and spurious parents. To filter out the spurious parents, the neural networks incorporate stochastic gating layers \citep{yamada2020feature} as the first layer, which serve as a variable selection mechanism. 
Stochastic gates $z_i$ are continuous relaxations of Bernoulli gates, utilizing clipped and mean-shifted Gaussian random variables:
\begin{equation}
   {z_i = \max{(0, \min{(1, \alpha_i + \varepsilon_i)})}, \quad \varepsilon_i \sim \mathcal{N}(0, \sigma_z^2)},
\label{eq:stg}
\end{equation}
where {$\alpha_i$} are the mean parameters for each dimension, $\varepsilon_i$ are the noise terms introduced for stochasticity, and $\sigma^2_z$ is the pre-defined noise variance. As the mean parameters are updated along with the neural network parameters, the values of $z_i$ are taken to be either 0 or non-zero (with a maximum value of 1). By taking the element-wise product of the sampled $z_i$ and the input, only the variables relevant for predicting the target are passed to the subsequent layers. 
Once training for network $f_i$ is complete, the stochastic noise is dropped ($\varepsilon_j = 0$). As formalized in line 20 of Algorithm \ref{alg:ocdm}, any candidate edge $(j, i)$ where the deterministic gate has collapsed to zero ($\alpha_j \le 0 \implies z_j = 0$) is soundly pruned from the initial graph edge set $\mathcal{E}^{\pi}$ to yield the final, sparse causal DAG $G^{\pi}_{*}$. This makes STG-NN a sparse regression model capable of performing variable selection. 

We believe that CAM-pruning and STG-NN pruning would perform similarly when the number of nodes in the causal structure is small. However, as the dataset grows in both dimensionality and sample size, STG-NN pruning is expected to outperform CAM-pruning due to its superior capability for nonlinear variable selection. Although the computational complexity of STG-NN pruning is higher than that of CAM-pruning ($\mathcal{O}(n \cdot E \cdot L \cdot m^2)$ vs. $\mathcal{O}(nk^2+k^3)$, where $n$ is the number of training samples, $E$ is the number of training epochs, $L$ is the number of neural network layers, $m$ is the number of neurons per layer, and $k$ is the number of spline basis functions), its practical execution can be greatly accelerated by standard training techniques such as GPU parallelization and mini-batch processing. Consequently, despite the scaling difference, the actual runtime of STG-NN pruning may be faster for high-dimensional or large-sample problems. Moreover, the computational inefficiency arising from variable-wise pruning may be alleviated by running the pruning procedure in parallel, as each variable is pruned independently. To further leverage the flexibility of STG-NNs, we empirically find that gradually increasing the regularization strength with the number of candidate parents improves pruning performance.

\section{Experiments}\label{sec:experiments}

\subsection{Data description}
We validated OCDM using data from simulated random graphs and pseudo-real data that emulate a real multistage process \citep{gobler2024textttcausalassembly}. The simulation processes for these datasets are described below.

\subsubsection{Simulation data}\label{sec:data_simulation}
We first tested OCDM on synthetic data generated from simulated random graphs with a nonlinear additive noise causal model (Eq. \eqref{eq:nonlinear_additive_noise}). The simulated graphs were designed to mimic multistage processes. The dataset was generated as follows. First, we generated $w$ random DAGs using the Erd\H{o}s-R\'enyi model \citep{erdios1959random} with $k$ nodes, setting the average number of edges for each DAG to be $k$ as well. Each of these random graphs represents a stage in a multistage dataset. Next, while maintaining the hierarchy between the stages (i.e., edges cannot exist from lower layers to upper layers), we randomly connected the graphs with a probability equal to that of intra-stage edges. Following previous methods \citep{buhlmann2014cam, rolland2022score, sanchez2023diffusion}, we sampled the link functions from a Gaussian process with a zero-mean function and a unit radial basis function covariance kernel. The noise distribution for each variable was set to Gaussian, with variance sampled from a uniform distribution $\mathcal{U}(0.4, 0.8)$. We tested two scenarios: a few-stage scenario ($w=3, k=10$), referred to as Simulation 1, and a many-stage scenario ($w=20, k=10$), referred to as Simulation 2. {We also used the dataset from Simulation 1 to investigate the impact of noisy stage information. Specifically, we introduced noise by selecting a proportion $p \in \{0.1, 0.2, 0.3, 0.4, 0.5\}$ of variables and randomly reassigning their stage labels to stages different from their original ones, while keeping the data values of all variables unchanged.}

{Furthermore, we generated additional datasets under more challenging scenarios designed to reflect characteristics commonly observed in real-world data. Based on the same randomly generated DAG used in  Simulation 1 with $w=3$ and $k=10$, we generated mixed-type data in which 30\% of the variables were binary. This dataset does not conform to the causal model assumed by OCDM and the other baseline methods considered (Eq. \eqref{eq:nonlinear_additive_noise}), rendering causal discovery using these methods theoretically unsuitable. However, because real-world datasets often contain both continuous- and discrete-valued variables, conducting exploratory experiments beyond datasets that strictly follow the assumed causal model enables us to evaluate the robustness of the proposed method under scenarios that deviate from the stated assumptions. Continuous-valued variables were generated using the same Gaussian process described above, while binary variables were generated by passing the outputs of the Gaussian process functions through a logistic function $\frac{1}{1+e^{-x}}$, and rounding the results to the nearest integer (0 or 1).}

\subsubsection{Pseudo-real data}
\begin{figure*}[ht]
    \centering
    \includegraphics[width=0.75\textwidth]{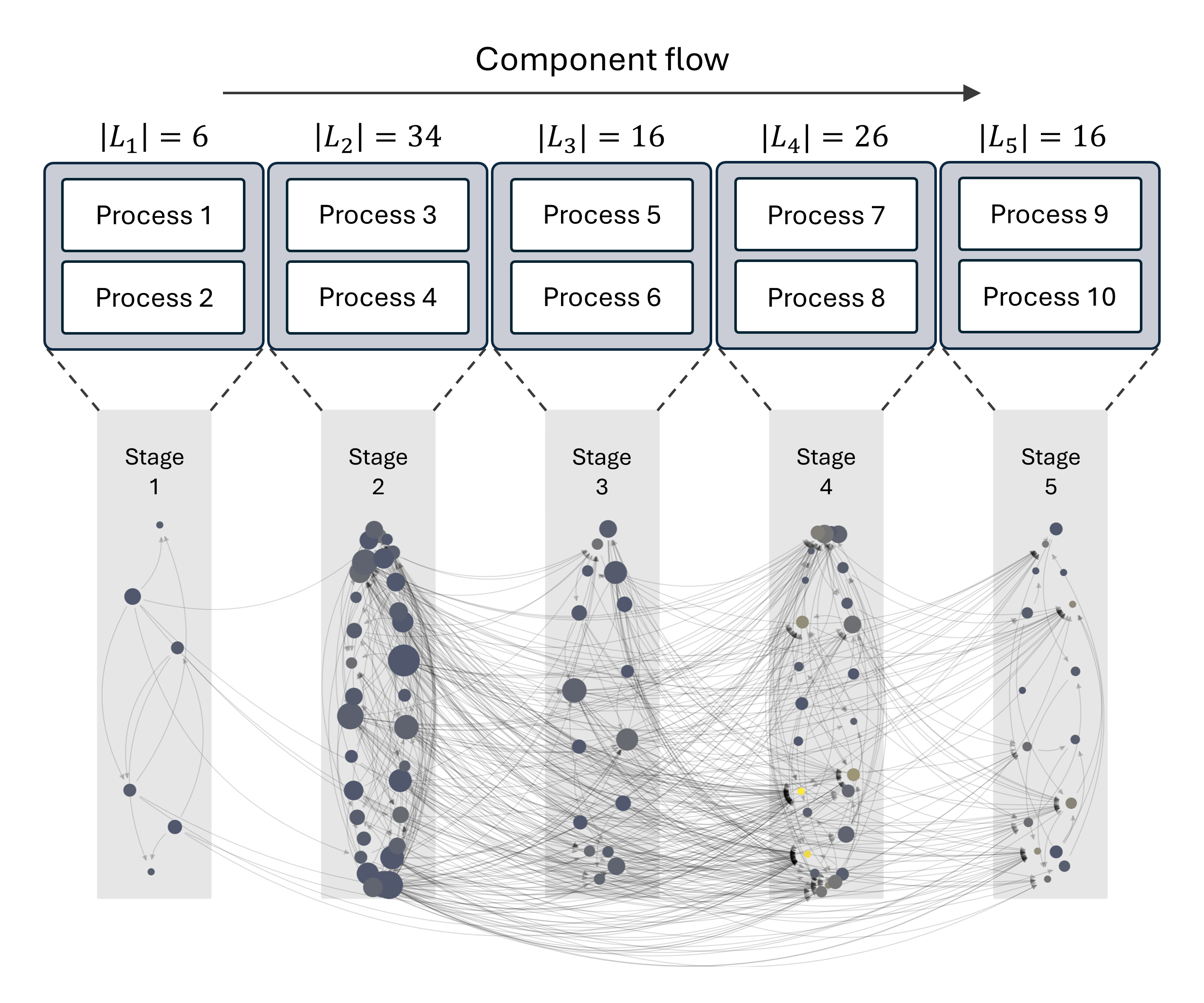}
    \caption{Process line of \emph{causalAssembly} and its ground-truth DAG (figure adapted from Figures 2 and 3 of \citep{gobler2024textttcausalassembly}). Each stage consists of nodes from two distinct adjacent processes. The size of each node represents the number of its out-edges, while the brightness indicates the number of its in-edges. $|L_i|$: number of nodes in stage $L_i$.}
    \label{fig:causalAssembly}
\end{figure*}

Obtaining the full causal structure of real systems, along with their corresponding data observations, is a challenging task. To overcome this, we utilized \emph{causalAssembly} \citep{gobler2024textttcausalassembly} to generate pseudo-real data for validating our methods on real-world-like data. CausalAssembly models a real assembly line from an automated manufacturing plant, where 10 processes---each a press-in or staking process---are performed in series. These processes are grouped into 5 stages, creating a multistage manufacturing process with a 5-stage layered DAG structure. 
A total of 98 variables are collected across the 5 stages, with the number of variables per stage ranging from 6 to 34. Fig. \ref{fig:causalAssembly} provides a detailed view of the causalAssembly setup. 
The causal relationships among the variables within the modeled process are defined by domain experts based on their knowledge of the physical process, making causalAssembly a near-real benchmark for evaluating causal discovery methods. We used the Python package provided by \citet{gobler2024textttcausalassembly} and followed their outlined data generation process to obtain samples from the causalAssembly model. 

All experiments were performed on a Windows workstation with an Intel Core i5-11400 CPU, an NVIDIA RTX 3070 GPU, and 64GB of DDR4 RAM. We perform 10 repeated trials for each method in all experiments.

\subsection{Causal order search results}
We begin our analysis by evaluating the order search performance of order-based causal discovery methods. We compared our method with four previously proposed order-search methods: CAM \citep{buhlmann2014cam}, SCORE \citep{rolland2022score}, DAS \citep{montagna2023scalable}, and the vanilla version of DiffAN \citep{sanchez2023diffusion}. We do not present the order inference results of CAM for Simulation 2, as it failed to terminate within a reasonable time period (24 hours).

Because order-based methods first learn the causal order before constructing the causal graph, the quality of the inferred causal order is a critical prerequisite for obtaining a high-quality estimate of the causal graph. A pruned DAG cannot introduce an edge that does not already exist in the order-connected graph. Therefore, we evaluated the order-based methods, including OCDM, with respect to their order-connected graphs through the metric of topological order divergence \citep{rolland2022score}, which is defined as follows:
\begin{equation}
D_{top}(\pi, \mathbf{A}) = \sum_{i=1}^d \sum_{j:\pi_i > \pi_j} \mathbf{A}_{ij}, 
\end{equation}
where $\pi$ represents a causal order, and $\mathbf{A}$ is the target adjacency matrix. $D_{top}(\pi, \mathbf{A})$ counts the number of edges in the true graph that cannot be recovered using the order-connected graph generated by $\pi$. In other words, assuming an edge is drawn from a preceding variable to all succeeding variables given some causal order $\pi$, $D_{top}(\pi, \mathbf{A})$ counts the edges that cannot exist in the DAG $\mathbf{A}$ because the causal direction between a pair of variables was wrongly inferred due to an incorrect causal order.
Hence, $D_{top}(\pi, \mathbf{A})$ provides a measure of the accuracy of the inferred causal order, independent of the pruning method.  

\begin{figure}[ht]
    \centering
    \includegraphics[width=\columnwidth]{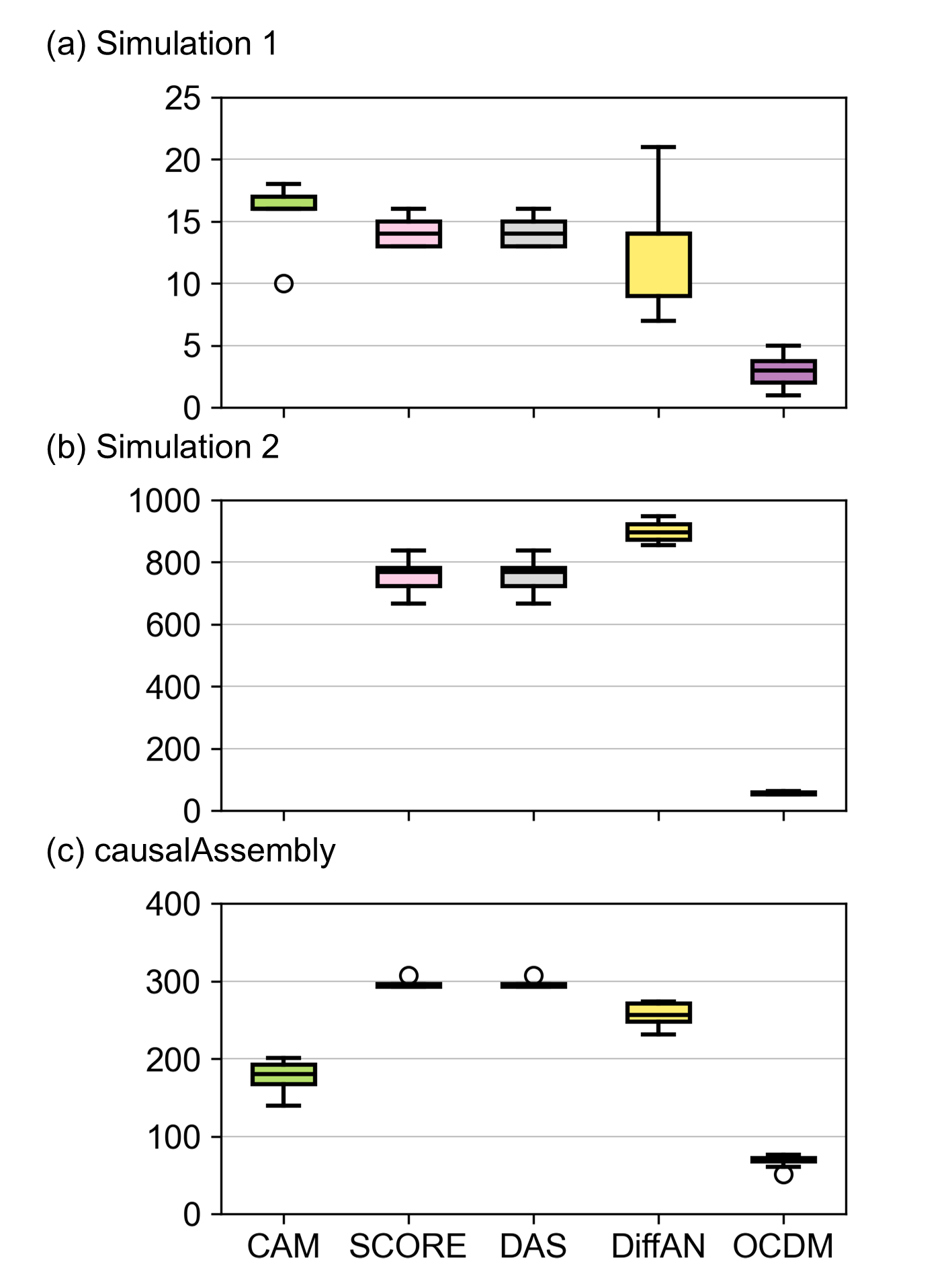}
    \caption{Topological order divergence ($D_{top}$) on order-connected graphs. Lower values are preferred, indicating a more accurate causal order.}
    \label{fig:top_d}
\end{figure}

Fig. \ref{fig:top_d} presents the order search results for the datasets considered in this study. Across all datasets, the OCDM order search algorithm achieved the best estimate of causal order, outperforming previously proposed order-based causal discovery methods in terms of topological order divergence. This indicates that OCDM infers the causal order that generates order-connected graphs with the least number of edges that do not exist in the ground-truth DAG, resulting in the most accurate causal structure prior to pruning. Although all methods performed decently on the low-dimensional dataset (Simulation 1) with a low number of false edges, OCDM estimated a highly accurate causal order with a mean $D_{top}=3$, which is 67\% lower than that of the best-performing competing method, DiffAN. Furthermore, while the quality of the inferred causal order degraded rapidly as the dimensionality of the dataset increased, OCDM still estimated the most accurate causal order, achieving a 90\% lower $D_{top}$ than SCORE, which performed the best among the competing methods. Additionally, OCDM demonstrated the best $D_{top}$ for the pseudo-real dataset, which is relatively high-dimensional and contains potentially more complex causal relationships. Therefore, the utilization of multistage information by searching for the causal order within each stage in reverse order contributed to a more accurate causal order, as expected.

\subsection{Causal DAG estimation results}
\begin{figure*}[ht]
    \centering
    \includegraphics[width=\textwidth]{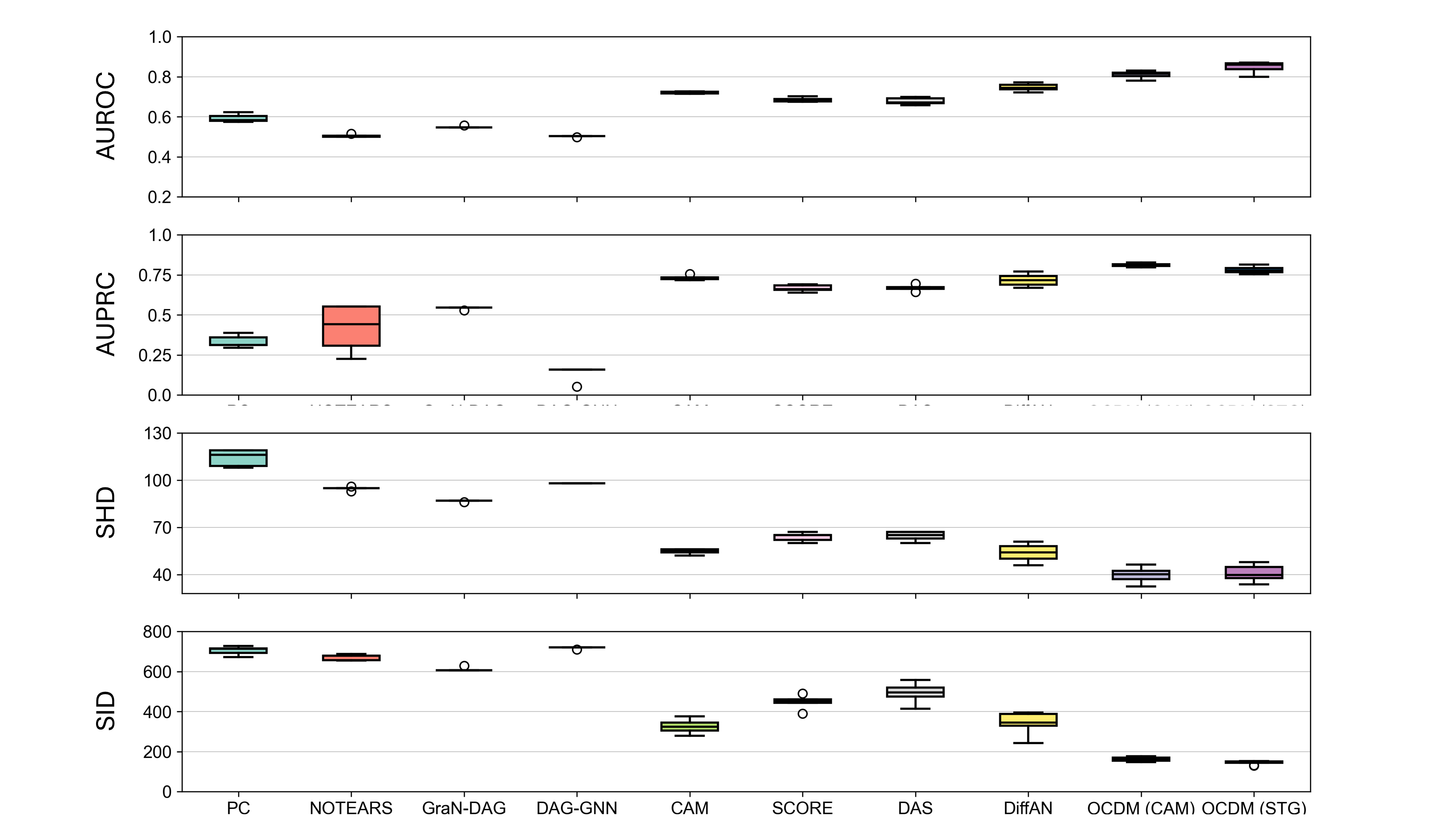}
    \caption{Causal discovery results for Simulation 1 ($w=3, k=10$). Higher values indicate better performance for AUROC and AUPRC, while lower values are preferred for SHD and SID.}
    \label{fig:simulation1}
\end{figure*}

\begin{figure*}[ht]
    \centering
    \includegraphics[width=\textwidth]{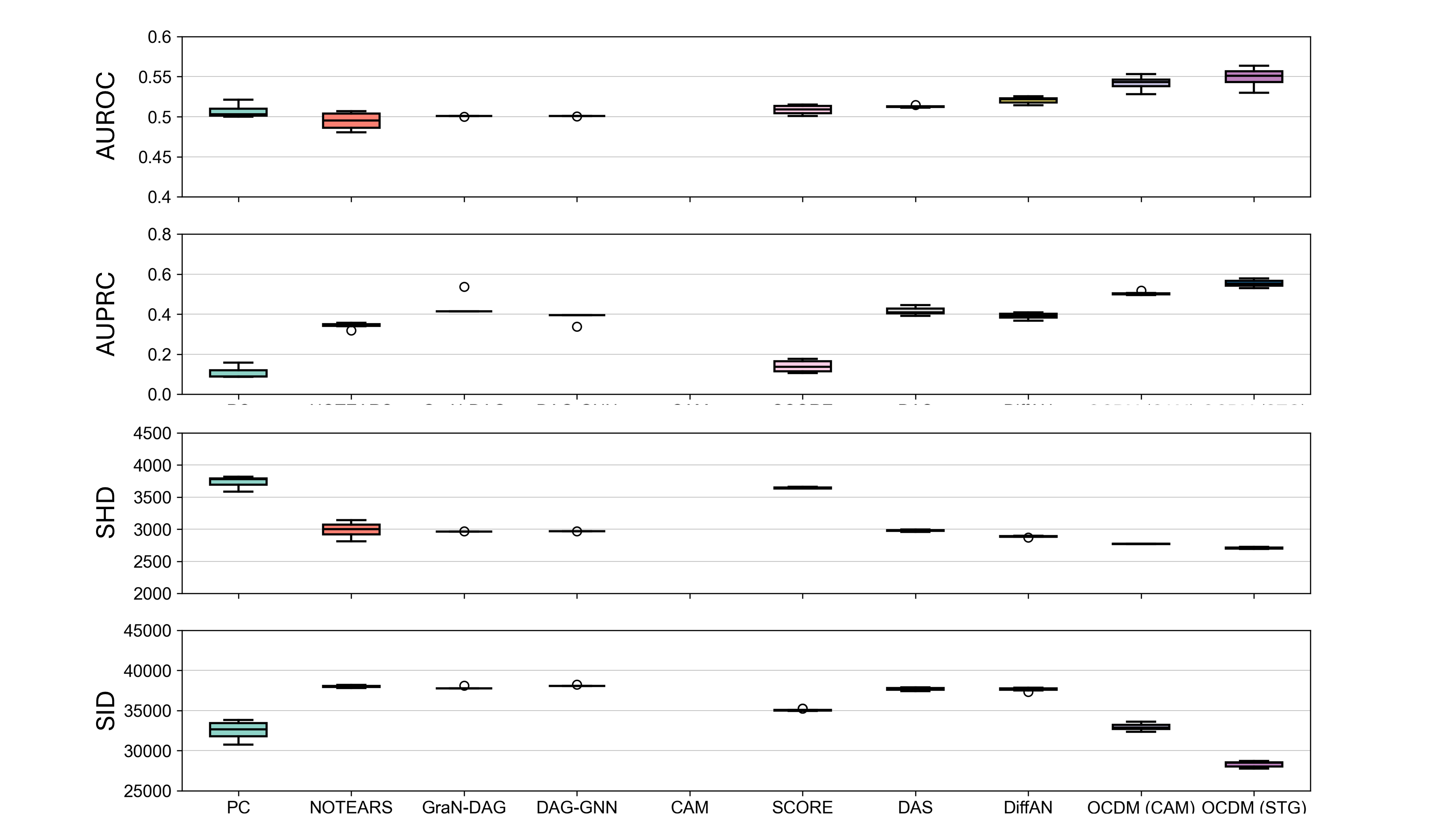}
    \caption{Causal discovery results for Simulation 2 ($w=20, k=10$). Higher values indicate better performance for AUROC and AUPRC, while lower values are preferred for SHD and SID.}
    \label{fig:simulation2}
\end{figure*}

\begin{figure*}[ht]
    \centering
    \includegraphics[width=\textwidth]{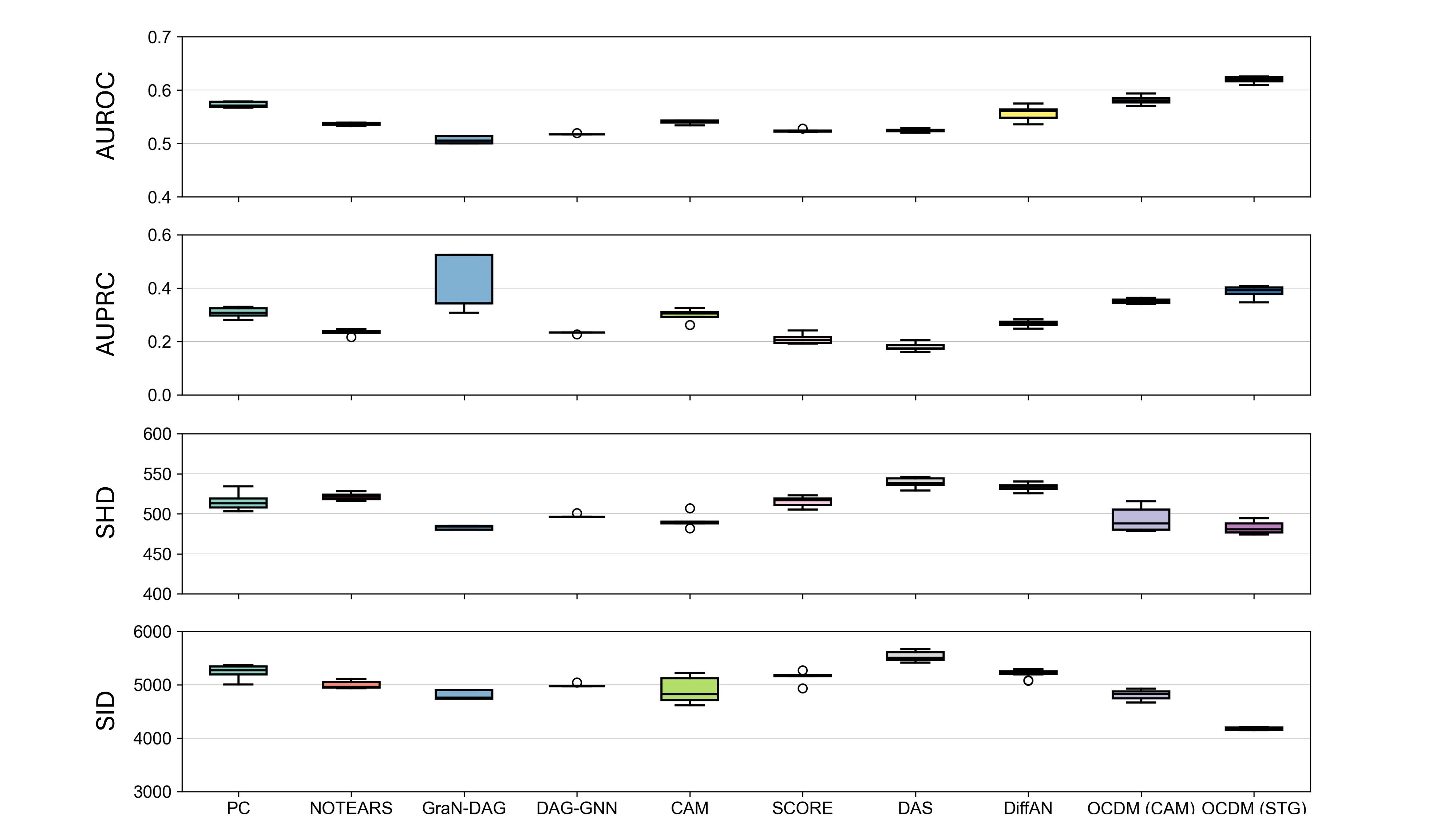}
    \caption{Causal discovery results on the pseudo-real data (\emph{causalAssembly}). Higher values indicate better performance for AUROC and AUPRC, while lower values are preferred for SHD and SID.}
    \label{fig:causalassembly}
\end{figure*}

OCDM generates an order-connected graph based on the inferred causal order, which is then pruned to produce the final causal structure estimate. To evaluate the performance of OCDM, we present the results for both the conventional CAM-pruned DAG (denoted as OCDM (CAM)) and the proposed STG-NN pruned DAG (denoted as OCDM (STG)) in Fig. \ref{fig:simulation1}, \ref{fig:simulation2}, and 
 \ref{fig:causalassembly}, corresponding to Simulation 1, Simulation 2, and pseudo-real data, respectively.
We compared OCDM to existing methods for causal discovery, including constraint- and score-based methods (PC-algorithm \citep{spirtes1991algorithm}, NOTEARS-MLP \citep{zheng2020learning}, DAG-GNN \citep{yu2019dag}, and GraN-DAG \footnote{We follow the pre- and post-processing steps provided in \citet{lachapelle2019gradient}} \citep{lachapelle2019gradient}) as well as order-based methods (CAM \citep{buhlmann2014cam}, SCORE \citep{rolland2022score}, DAS \citep{montagna2023scalable}, and DiffAN \citep{sanchez2023diffusion}). 
Details on the implementation scheme and hyperparameter configurations for each method can be found in Section II of the Supplementary Materials.

Because DAGs can be represented as binary matrices in the form of adjacency matrices, we evaluated the methods using the area under the receiver operating characteristics curve (AUROC) and the area under the precision-recall curve (AUPRC). Additionally, we evaluated the methods using structural Hamming distance (SHD) and structural intervention distance (SID) \citep{peters2015structural}, which are graph-specific evaluation criteria. SHD measures the similarity between two DAGs based on the number of inconsistent edges, while SID measures the similarity based on the number of inconsistent intervention distributions. The detailed definitions of all metrics are provided in Section III of the Supplementary Materials.

\subsubsection{Results on simulation data}
We first evaluated our proposed method and the baseline methods on the simulated datasets. Fig. \ref{fig:simulation1} presents the causal discovery performance for the few-stage scenario (3 stages, 10 nodes per stage). Most methods performed reasonably well, given the low-dimensional structure of the tested simulation data. However, regardless of the pruning method, OCDM consistently outperformed all competing methods across all metrics, demonstrating the effectiveness of accurately inferring the causal order before constructing the causal DAG. By selecting parents from a candidate pool based on the inferred causal order, OCDM reduces the risk of overlooking true parents. 

The two pruning methods under consideration, conventional CAM-pruning and the proposed STG-NN pruning, showed comparable performance across all metrics. 
STG-NN pruning performed better than CAM-pruning with respect to AUROC and SID, while CAM-pruning performed better in terms of AUPRC and SHD. However, the differences between the two pruning methods were marginal across all considered metrics. We speculate that this is because spline regression functions can fit the data as well as neural networks in low-dimensional settings, such as with a maximum dimensionality of $d=29$ (29 candidate parents for the leaf node in a causal order of 30 variables). 

Fig. \ref{fig:simulation2} presents the causal discovery performance of the considered methods in the many-stage scenario (20 stages, 10 nodes per stage). Again, we do not show the results of CAM, as it failed to train within a reasonable time period.
Although the performance of all methods decreased sharply, OCDM consistently outperformed the others across all metrics.

The strengths of OCDM are also highlighted in its computational efficiency as the size of the dataset grows. The total time taken by OCDM (STG-NN pruned) to infer the causal DAG of 200 nodes was approximately 2.75 hours, which was only slightly slower than the PC-algorithm, which took approximately 2.1 hours to train. However, the causal structure inferred by the PC-algorithm was far less accurate than that learned by OCDM. The time taken by other methods considered in this study to successfully infer a causal DAG of 200 nodes ranged from approximately 4 hours (DAG-GNN) to more than 12 hours (SCORE). Therefore, OCDM displayed the best causal discovery performance while consuming significantly less time to train. 

The STG-NN pruned OCDM outperformed the CAM-pruned OCDM across all metrics considered in this scenario of 20 stages and 10 nodes per stage, but spent only half the time required to prune all nodes via CAM-pruning. Specifically, STG-NN pruning took approximately 1.2 hours, whereas CAM-pruning took approximately 2.2 hours. 
The improvements in both DAG accuracy and computational efficiency are likely attributed to the increased dimensionality of the data. As the overall dimensionality increases and more variables are introduced, the CAM-pruning method may struggle to accurately select true parents from the candidate parent set through variable selection. In contrast, neural networks can more flexibly learn predictive functions regardless of input size. Thus, combined with the stochastic gating mechanism for variable selection, the STG-NN pruned OCDM is expected to yield superior causal discovery results in high-dimensional data compared to the CAM-pruned methods. This highlights the importance of applying a pruning technique that is appropriate for the typically high-dimensional multistage process data. In conclusion, the experimental results on the higher-dimensional simulation data demonstrate that OCDM has advantages over other methods in both order search and pruning, leading to a more accurate discovery of causal relationships among the variables in the dataset. 

\subsubsection{{Results on data with noisy stage information}}
\begin{figure*}[ht]
    \centering
    \includegraphics[width=\textwidth]{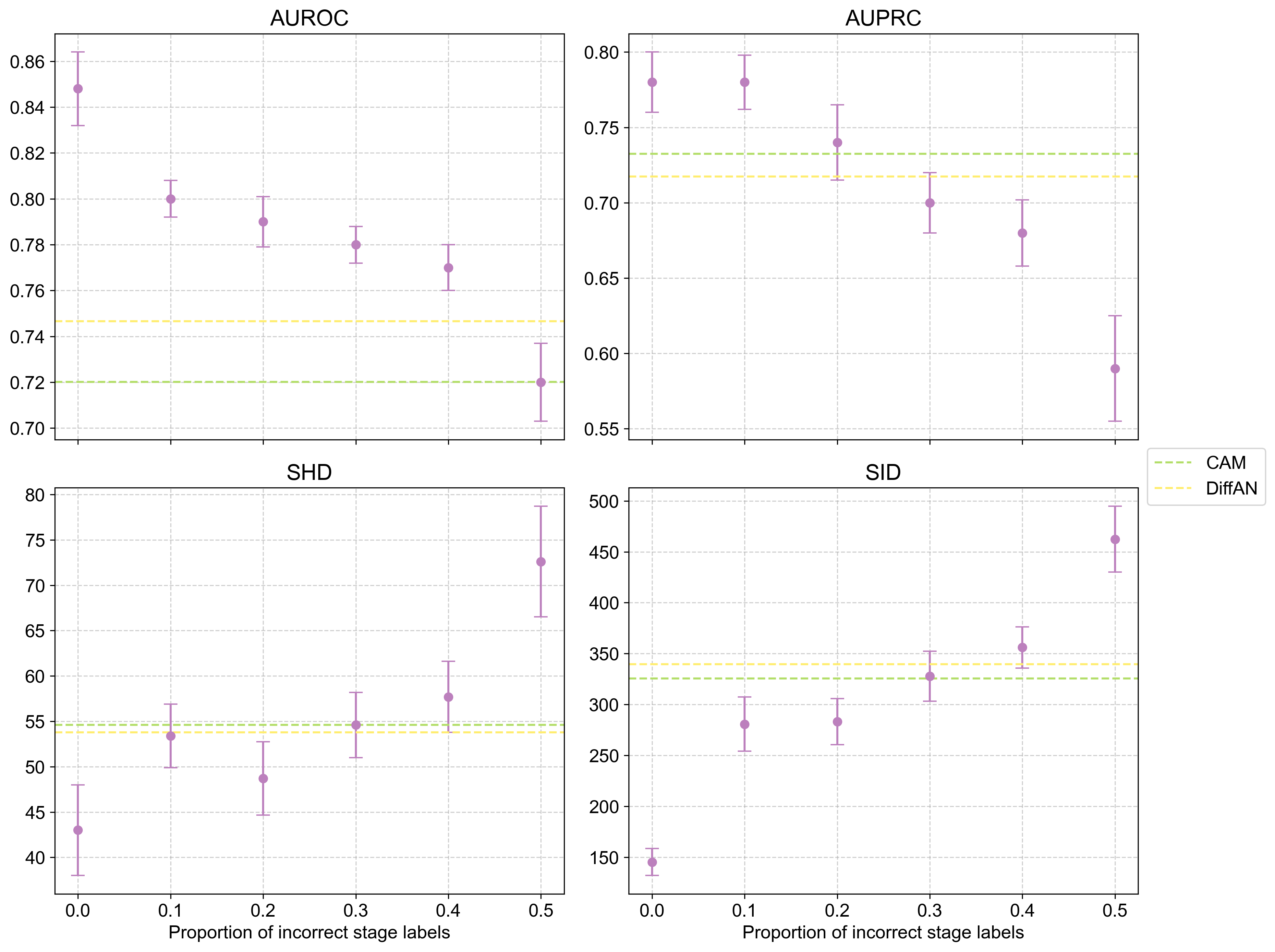}
    \caption{{Causal discovery results for Simulation 1 ($w=3, k=10$) under noisy stage information. The results of the best-performing baseline methods, CAM and DiffAN, are depicted with green and yellow dotted lines, respectively. Higher values indicate better performance for AUROC and AUPRC, while lower values are preferred for SHD and SID.}}
    \label{fig:noisy}
\end{figure*}

{Although OCDM leverages stage information to enhance causal order search and DAG inference, in practice, such information may not always be accurately available. To evaluate the robustness of OCDM under noisy stage information, we conducted experiments in which a proportion $p \in \{0.1, 0.2, 0.3, 0.4, 0.5\}$ of variables were randomly assigned incorrect stage labels. Fig. \ref{fig:noisy} presents the performance of OCDM under varying noise levels, where $p=0$ corresponds to the baseline case with perfectly accurate stage information. The results of the best-performing baseline methods, CAM and DiffAN, are shown as green and yellow dotted lines, respectively.}

{As expected, OCDM exhibited a gradual degradation across all evaluation metrics with increasing noise levels---showing lower AUROC and AUPRC and higher SHD and SID. This trend reflects the impact of inaccurate prior information, which leads to incorrect causal ordering for mislabeled variables. Nevertheless, OCDM consistently outperformed the baseline methods for noise levels below $p=0.2$ and remained comparable to them at $p=0.3$. These results demonstrate that the robustness of OCDM; even when stage information is partially incorrect, it remains beneficial for causal structure inference.}

{The observed robustness of OCDM can be attributed to two factors. First, in the causal order inference phase, which leverages the Jacobian of the score function, mislabeled variables are not placed arbitrarily but are typically positioned near the boundaries of their incorrectly assigned stages. For example, a variable originating from an earlier stage tends to appear at the beginning of its new, incorrect stage, while a variable from a later stage is positioned near the end. This boundary alignment minimizes the number of incorrect edges in the order-connected graph constructed from the inferred causal order. Second, the pruning procedure can serve as an error-correction mechanism: during this phase, the order-connected graph is refined into the final DAG by removing edges lacking sufficient evidence of causality. Edges incorrectly introduced due to noisy stage labels tend to be weakly predictive of their supposed child variables and are thus likely to be pruned. This further mitigates the impact of noisy stage information.}

{Despite OCDM's robustness to noisy stage information, its performance fell below that of the baselines at extremely high noisy levels ($p=0.4$ or $0.5$), as the erroneous stage information increasingly forces an incorrect causal order regardless of the Jacobian-based evidence. Therefore, OCDM is most effective when stage information is available with at least moderate reliability and is not recommended when the variable origins are highly uncertain. We further speculate that OCDM would demonstrate comparable or even greater robustness under partially known stage information, as partial labeling is less detrimental than incorrect prior knowledge.}

\subsubsection{{Results on mixed-type data}}
{All methods considered in this study, except for the PC-algorithm \citep{spirtes1991algorithm}, assume that the data are generated from a nonlinear additive noise model (Eq. \eqref{eq:nonlinear_additive_noise}), and consequently that all variables are continuous. However, mixed-type datasets---containing both continuous and discrete variables---are commonly observed in real multistage systems. To investigate OCDM’s performance in such cases, we generated a mixed-type dataset based on the layered DAG used in Simulation 1. Continuous variables were generated using Gaussian process samples, while binary variables were obtained by applying the logistic function to the outputs of Gaussian process functions (Sec. \ref{sec:data_simulation}).}

\begin{figure*}[ht]
    \centering
    \includegraphics[width=\textwidth]{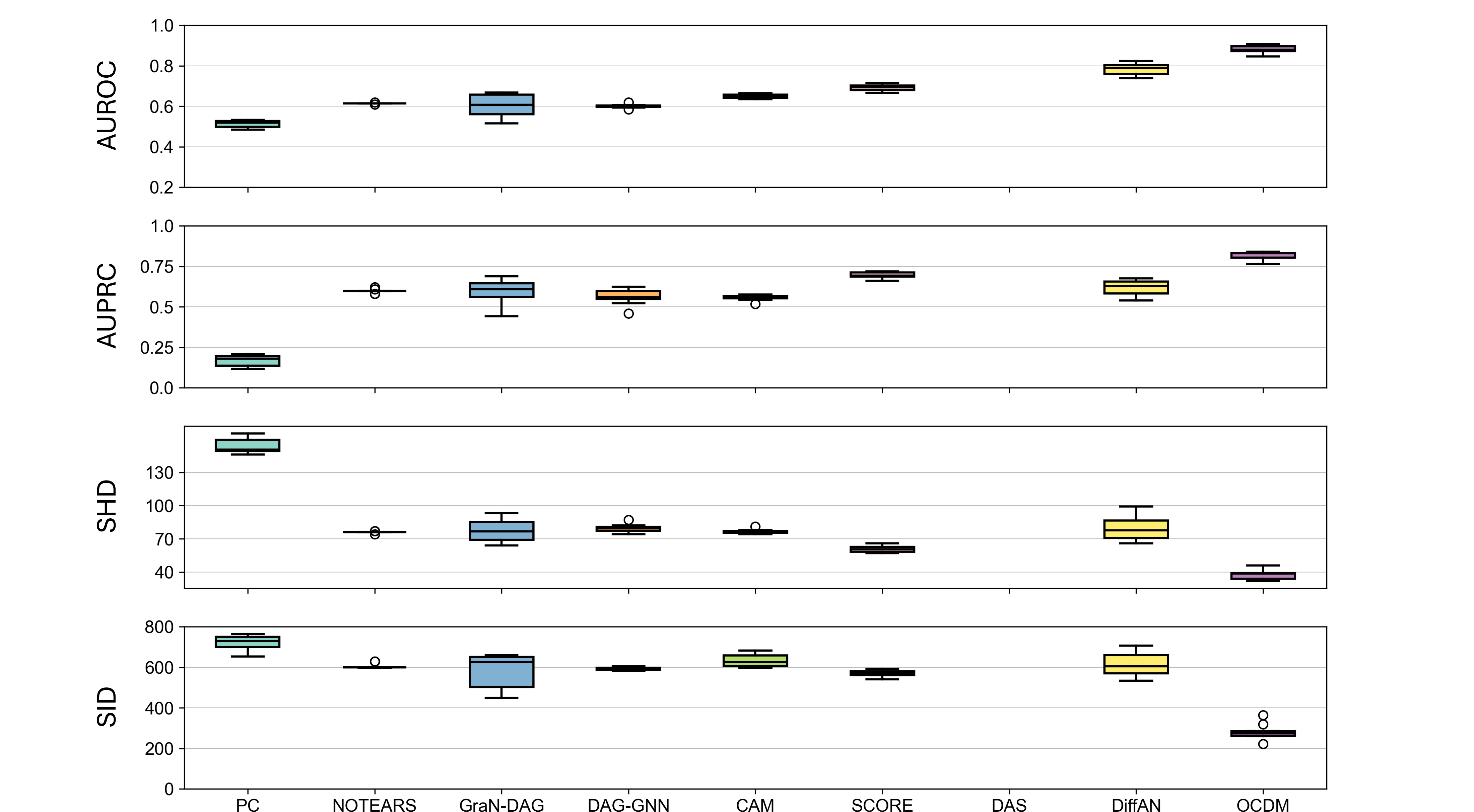}
    \caption{{Causal discovery results for mixed-type data with $w=3, k=10$. Higher values indicate better performance for AUROC and AUPRC, while lower values are preferred for SHD and SID.}}
    \label{fig:mixed}
\end{figure*}

{Fig. \ref{fig:mixed} presents the causal discovery results of OCDM and the baseline methods on the mixed-type dataset. Results for DAS \citep{montagna2023scalable} are omitted due to its instability when trained on data with discrete variables, likely caused by numerical brittleness during matrix inversions. OCDM consistently outperformed all baseline methods across all considered metrics. Notably, the PC-algorithm---theoretically suitable for mixed-type data---performed poorly, particularly in SHD and SID. In contrast, OCDM accurately recovered the underlying causal graph despite relying on the nonlinear additive noise model assumption. 
We attribute these counterintuitive results to OCDM's STG-NN-based pruning phase. While the causal order search may be affected by the theoretical mismatch between the assumed causal model and the mixed-type data at hand, the pruning phase likely enhances OCDM's capability to handle mixed-type data, as classification tasks are generally easier to optimize than regression tasks. These empirical results highlight the robustness and practical effectiveness of OCDM for causal discovery in heterogeneous data settings, suggesting that OCDM remains a promising framework even when the observed data do not fully conform to the assumed generative model.}

\subsubsection{Results on pseudo-real data}
Lastly, we compared our proposed method to other causal discovery methods in terms of their ability to infer the causal structure of the causalAssembly data \citep{gobler2024textttcausalassembly}, which are pseudo-real data generated from a simulator of a real multistage manufacturing process. Experiments on the causalAssembly data allow us to assess how the proposed method performs on real-world multistage manufacturing data. Fig. \ref{fig:causalassembly} presents the causal discovery results for the methods applied to the causalAssembly data. 
Consistent with the results from the simulation data, OCDM (STG-NN pruned) outperformed the other methods across all performance metrics. 
While OCDM consistently performed the best across all metrics considered, its superiority was most pronounced in terms of SID. 
This indicates that the causal structure inferred using OCDM is not only closer to the ground-truth causal structure of causalAssembly compared to those inferred by other methods, but also more suitable for performing further causal inference tasks based on the inferred causal relationships.

Similar to the findings in Simulation 2, the superiority of STG-NN pruning over CAM-pruning was evident in the experiments with the causalAssembly data. 
We attribute this advantage to the STG-NN's ability to perform sparse regression and select appropriate variables from the pool of candidate parents, as observed in the simulation data experiments. We believe that the advantages of STG-NN in exploring a broader range of input combinations compared to the sparse spline regression employed in CAM-pruning will be more pronounced for higher-dimensional data with complex causal relationships between variables, making STG-NN an effective alternative pruning method.

\section{Conclusions and Limitations}
In this study, we introduced OCDM, an order-based causal discovery framework designed specifically for multistage processes. Recognizing that the order of stages and the origins of each variable are often known in multistage data, OCDM leverages this prior knowledge through a structural, knowledge-informed causal order search. The causal order is inferred within each stage and then concatenated to infer the causal order for the entire system. {Because the causal order of variables must conform to the sequence of stages, this approach enables OCDM to obtain a more accurate causal order than other order-based methods, which are often prone to inferring counterintuitive causal directions. The proposed order search algorithm could be further enhanced in a similar manner if finer-grained prior information, such as intra-stage subprocess details, is available.}

The inferred causal order is then used to generate an order-connected graph, which is a DAG in which edges connect each node to all succeeding nodes in the causal order. The generated order-connected graph undergoes pruning to eliminate spurious edges and yield a final estimate of the causal structure. We proposed the use of stochastic gated neural networks as an alternative to the conventional pruning method proposed in \citet{buhlmann2014cam}, enhancing the incorporation of multistage process characteristics. We evaluated the performance of OCDM on various datasets of varying complexity, including pseudo-real data that mimic a real multistage manufacturing process. The results demonstrated OCDM's superior ability to accurately infer the underlying causal structure compared to existing causal discovery methods. {Furthermore, OCDM exhibited substantial robustness across challenging scenarios often encountered in real systems, such as the presence of noisy stage information or mixed-type datasets that deviate from the study's assumptions. Importantly, these findings suggest an extended applicability of OCDM, potentially enabling its use even when stage information is unavailable but can be inferred with high confidence.}

Despite the suitability of OCDM for causal discovery in multistage process data, several limitations remain that could be addressed in future research. First, OCDM is based on score matching methods \citep{rolland2022score, sanchez2023diffusion} to infer the causal order, which is applicable only to nonlinear Gaussian (or location-scale family) additive noise models. Although this causal model is flexible enough to encompass a wide range of processes, developing causal discovery methods for multistage processes under different causal models would be a valuable avenue for exploration. Furthermore, these underlying assumptions restrict our method to settings without latent confounders, where iterative score matching-based discovery can fail to accurately identify leaf nodes under causal insufficiency. Consequently, adapting the stage-informed framework of OCDM to iterative discovery methods designed for latent confounding (e.g. AdaScore \citep{montagna2024score}) represents a promising direction to alleviate this limitation. Similarly, adapting deep cross-modal and representation-learning architectures from multimedia causal frameworks to highly heterogeneous, mixed-type industrial settings constitutes a promising avenue for future research. Additionally, the causal discovery methods considered in this study, including OCDM, were susceptible to over-pruning, resulting in the removal of true edges along with spurious ones. Although the improved quality of order-connected graphs and the proposed STG-NN pruning helped mitigate this issue in OCDM compared to other methods, the causal discovery results remained limited to identifying only a subset of the true edges. Hence, empirical strategies such as the use of ensemble methods or carefully designed thresholding schemes based on prior domain knowledge may help prevent over-pruning, depending on the characteristics of the dataset. Conversely, challenges arising during the pruning phase may stem from the fact that both pruning methods considered in the study---STG-NN pruning and CAM-pruning---lack theoretical guarantees for recovering the true parent set. Therefore, validating the proposed method across diverse real-world scenarios, as well as developing domain-specific empirical strategies to further enhance its performance, would be an interesting direction for future work. Lastly, although we attempted to validate the proposed method under various realistic scenarios, our experiments rely heavily on simulation data, particularly those with a moderate dimensionality of up to 200 variables. This limitation stems from practical challenges, such as the difficulty of  obtaining the true causal structure of real processes and the substantial computational resources required. Consequently, conducting experiments on datasets larger than those used in the study was infeasible within the specified computing environment (Sec. \ref{sec:experiments}). Therefore, validating the proposed method across diverse real-world scenarios, as well as developing domain-specific empirical strategies to further enhance its performance, would be an interesting direction for future work. Despite these limitations, OCDM constitutes a meaningful advancement over existing causal discovery methods by effectively leveraging the characteristics of multistage data within its discovery framework.

\section*{Acknowledgements}
This research was supported by the National Research Foundation of Korea (NRF) grant funded by the Korea government (MSIT) (2023R1A2C2005453, RS-2023-00218913).

\bibliographystyle{IEEEtranN}
\bibliography{bibliography}

\end{document}